# Performance Evaluation of Histogram Equalization and Fuzzy image Enhancement Techniques on Low Contrast Images

Ebele Onyedinma[1], Ikechukwu Onyenwe[2] and Hycinth Inyiama[3]

[1, 2] Department of Computer Science, Nnamdi Azikiwe University, Awka. Anambra State, Nigeria

[1]eg.osita@unizik.edu.ng, [2]ie.onyenwe@unizik.edu.ng, [3]hc.inyiama@unizik.edu.ng

**ABSTRACT**

Image enhancement aims at improving the information content of original image for a specific purpose. This purpose could be for visual interpretation or for effective extraction of required details. Nevertheless, some acquired images are often associated with pixels of low dynamic range and as such result in low contrast images. Enhancing the contrast therefore tends to increase the dynamic range of the gray levels in the acquired image so as to span the full intensity range. Techniques such as Histogram Equalization (HE) and fuzzy technique can be adopted for contrast enhancement. HE adjusts the contrast of an input image by modifying the intensity distribution of its histogram. It is characterized by providing a global approach to image enhancement, computationally fast and easy to implement approach but can introduce unnatural artifacts and other undesirable elements to the resulting image. Fuzzy technique on its part enhances image by mapping the image gray level intensities into a fuzzy plane using membership functions; modifying the membership functions as desired and mapping back into the gray level plane. Thus, details at desired areas can be enhanced at the expense of increase in computational cost. This paper explores the effect of the use of HE and fuzzy technique to enhance low contrast images. Their performances are evaluated using the Mean squared error (MSE), Peak to signal noise ratio (PSNR), entropy and Absolute mean brightness error (AMBE).

## 1. INTRODUCTION

Image enhancement is the process of manipulating an image so that the result is more suitable than the original for a specific application [1]. It aims at improving the quality and the information content of original image for either preprocessing or post processing operations. Image enhancement is by nature problem oriented; showing that the technique to be adopted is dependent on the need for the enhancement. It could be for visual interpretation as in medical imaging where the visual quality of an image is paramount or for machine perception where easily quantified image is considered most ideal. In either case, the objective is to provide a more context-efficient resultant image than the original. Common enhancement practices include image filtering, image smoothing, image sharpening and contrast enhancement.

Image contrast is the difference in intensity between the highest and lowest intensity levels in an image [1]. An image having a good number of pixels with high dynamic range is expected to have a high contrast while that with low dynamic range will be of low contrast. A low-contrast image can result from poor illumination, lack of dynamic range in the imaging sensor, or even wrong setting of the image acquisition's lens aperture. Contrast enhancement therefore, tends to increase the dynamic range of the gray levels in the image being processed so as to span the full intensity range. This enhancement technique can be used as a preprocessing step in speech recognition, texture synthesis, comparison of image processing software, intelligent transportation systems, computer graphics, digitizing, multidimensional systems, military, remote sensing, medical imaging, industrial production among others [2]. Generally, to improve contrast in digital images, histogram equalization (HE) is commonly used. Histogram of an image is the graphical representation of the distribution of pixels in an image at each intensity value of the image [3]. The components of an image histogram show in great detail characteristics exhibited by an image. Low contrast image for example, typically has narrow histogram located towards the middle of the intensity scale. Histogram equalization therefore is an act of adjusting the contrast of an image by modifying the intensity distribution of the histogram. Its objective is to give a linear trend to the cumulative probability function associated to the image thereby, resulting in an image with a uniform histogram. With this uniform histogram, an overall enhancement of the image is achieved.

Histogram Equalization has been a very effective and most widely used method to enhance the contrast of images. It is computationally fast and easy to implement. Despite its success in contrast enhancement, it fails in situations where it is necessary to preserve the brightest factor of an image. It gives un-natural artifacts like intensity saturation, over-enhancement and noise amplification which are not desirable especially to consumer electronic products [1][4]. In order to enhance contrast and preserve brightness, techniques that can decompose the input image into several sub-images, and then apply the classical HE process were adopted by many researchers. In the work of [5], Bi-histogram equalization (BBHE) was proposed. This forms two separate histograms from the same image and then equalizes them independently. The first one is the histogram of intensities that are less than the mean intensity while the second is the histogram of intensities that are greater than the mean intensity. BBHE actually reduces the mean brightness variation but it cannot solve enhancement problem effectively as it can result in unnatural enhancement in some cases and requires higher degree of preservation. An extension of BBHE referred to as minimum mean brightness error bi-histogram equalization (MMBEBHE) was proposed by [6] to provide maximum brightness preservation. They performed separation of the input image's histogram based on the threshold level as opposed to the input mean of BBHE. This actually minimizes the difference between input and output image's mean and preserves brightness better than BBHE, though its major drawback is high computational cost.

Nevertheless, these techniques offer global approach to image enhancement and cannot be adapted to local brightness features of the input image because only global histogram information over the whole image is used [7]. To enhance details over a small area, fuzzy enhancement technique can be adopted. In the work of [8], use of parametric fuzzy transform to enhance low contrast image was proposed. The fuzzy components of the original image were generated using parametric fuzzy partition and modified by fuzzy probabilistic operators. in another development, [9] opined that the integration of multiple fuzzy membership functions will provide an optimal contrast enhancement. In general, unlike conventional image enhancement method which is usually associated with losing local brightness details in the highly dark and bright areas; fuzzy technique tends to avoid discontinuity of gray values by modifying the gray values with the help of membership function. In fact, it has the ability to manage the imprecision encountered in images effectively. In the fuzzy technique, the image is considered as a range of fuzzy singletons having a membership value. These membership values represent the degree of image property (membership) in the range; the nearer the value to unity, the higher the membership grade. Conversely, the closer the value to zero, the lower the membership grade [1]. Fuzzy image enhancement is achieved by mapping image gray level intensities into a fuzzy plane using membership functions, the membership functions are modified for desired purpose, and the fuzzy plane is mapped back to image gray level intensities. The result is producing an image with enhanced details at desired portions of the image

## 2. MATERIALS AND METHOD

2.1 Histogram equalization

Given a digital image $f$, the probability of occurrence of intensity level $r_k$ in the image is given as:

$$p_r(r_k) = \frac{n_k}{N} \; ; \qquad k = 0,1,2,\dots,L-1 \qquad (1)$$

Where $N$ is the total number of pixels in the image; $n_k$ is the number of pixels that have intensity $r_k$ and $L$ is the number of possible intensity levels in the image. It should be noted that $p_r(r_k)$ relates with the histogram of input image; as a plot of $p_r(r_k)$ versus $r_k$ gives the histogram of input image. Performing a discrete transformation on the cumulative distribution function CDF of the input image as follows

$$s_k = T(r_k) = (L-1) \sum_{j=0}^{k} p_r(r_j) \qquad (2)$$

Note that the cumulative distribution function corresponding to $p_r$ is given by

$$\sum_{j=0}^{k} p_r(r_j) \qquad (3)$$

Thus, the intensity transformation will produce

$$\frac{(L-1)}{N} \sum_{j=0}^{k} n_j \qquad k = 0,1,2,3,\dots,L-1 \qquad (4)$$

By mapping each pixel in the input image with intensity $r_k$ into a corresponding pixel with level $s_k$ in the output image produces an equalized image.

2.2 Fuzzy Image Enhancement Technique

Image enhancement using fuzzy technique involves three major steps:
    i. Image Fuzzification
    ii. Membership Function Modification
    iii. Image Defuzzification

The general structure of the fuzzy image processing is depicted in figure 1.

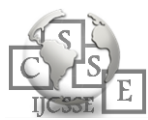

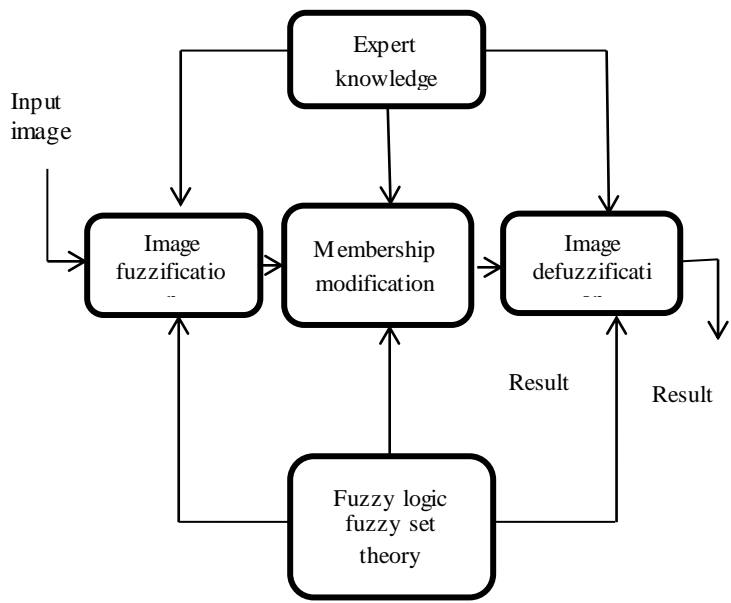

*Fig. 1. Steps in Fuzzy image processing*

**Image fuzzification**: This is the mapping of each scalar input by a corresponding fuzzy value based on the applicable membership function. In other words, it is the transformation of the image data from the gray level plane to the membership plane.

**Membership modification:** Having transformed the image data from gray-level plane to the membership plane, next step is to apply appropriate fuzzy techniques to modify the membership values. This can be a fuzzy clustering, a fuzzy rule-based approach, fuzzy classification approach, a fuzzy aggregation approach among others. In this context, fuzzy rule-based approach was adopted: Firstly, the outputs of all the parts of an antecedent are combined to produce a single value using max or min operation. An implication method is then applied to the single output of the antecedent of each rule to provide a corresponding output to that rule. Finally, aggregation method is applied to the fuzzy sets from the implication method to yield a single output fuzzy set.

**Deffuzification:** Outputs from the membership modifications are always fuzzy. To obtain a crisp value therefore, there is need to reverse the process of fuzzification (defuzzification) by applying any of the defuzzy methods such as center of area , inverse membership function and mean of maximum depending on the fuzzy approach adopted[1][10]. Thus, by computing the center of gravity of the aggregated fuzzy set, a crisp result is achieved.

## 2.3 Description of performance tools

The following image measuring tools: Absolute Mean Brightness Error (AMBE), Mean Squared Error (MSE), Peak to Signal Noise Ratio (PSNR) and Entropy were applied to both the HE and Fuzzy enhanced images on two low contrast images (figure2 and figure 3) to ascertain the quality of enhanced image against the original images.

**The Mean squared error (MSE)**: this represents the cumulative squared error between the enhanced image and the original image; the lower the value of MSE, the lower the error. MSE is given as

$$\frac{1}{MN}\sum_{M}^{i=0}\sum_{N}^{j=1}(f(i,j) - f'(i,j))^2 \qquad (5)$$

In the above equation, $M$ and $I$ represent the number of rows and columns in the input images with index i and j respectively. $f(i,j)$ represents the original image at location (i, j) and $f'(i,j)$ represents the degraded image at location (i,j).

**Absolute mean brightness error (AMBE)** which is defined as the absolute difference between the input and the output image mean. The expression is given as:

$$AMBE = |E(X) - E(Y) \qquad (6)$$

Where E (X) is the mean of the input image and E (Y) is the mean of the output image. Lower AMBE indicates that the brightness is better preserved.

**Entropy:** Entropy is a well-known statistical measure of randomness that can be used to characterize the texture of the input image. It measures the richness of the details in the output image. Entropy is defined as

$$-\sum_{i=1}^{n} p_i \log_2 p_i \qquad (7)$$

$p_i$ value is the occurrence probability of a given pixel. Higher entropy signifies higher image details preservation.

**PSNR:** Peak to signal noise ratio defines the ratio between the maximum possible power of a signal and the power of corrupting noise that affects the quality of image. PSNR represents a measure of the peak error.

$$PSNR = 10 \log_{10} \frac{(L-1)^2}{MSE} \qquad (8)$$

Where MSE is the mean squared error and L is the number of discrete gray levels. The greater the PSNR value, the better the output image quality. That is, higher value of PSNR indicates that the reconstruction is of higher quality.

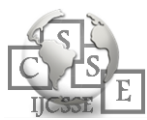

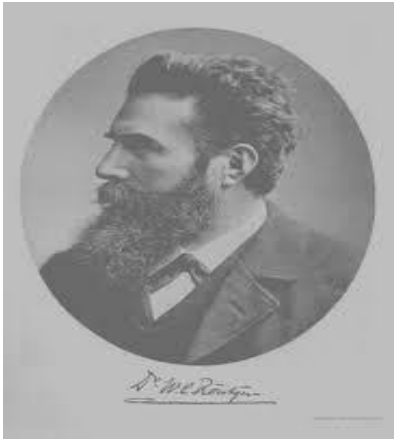

*Fig. 2. Low contrast image1*

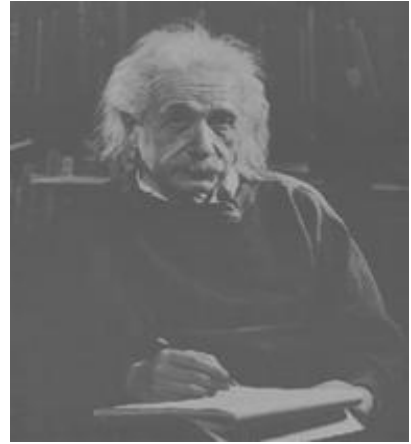

*Fig. 3. Low contrast image2*

## 3. RESULTS AND DISCUSSION

This section presents the performance of HE and fuzzy algorithms on low contrast images using the qualitative performance measures stated in section 2.3. The performance evaluation was carried out in Matlab 7.1b. Table 1 shows data obtained from the evaluation.

*Table 1: Summary of results*

|  |  | *MSE* | *PSNR* | *Entropy* | *AMBE* |
|---|---|---|---|---|---|
| Image 1 | HE | 2813.79 | 13.67 | 4.7082 | 30.4847 |
|  | Fuzzy | 2511.28 | 14.17 | 4.7935 | 14.7786 |
| Image 2 | HE | 4398.53 | 11.73 | 4.1760 | 25.1072 |
|  | Fuzzy | 2486.88 | 14.21 | 4.5128 | 38.6297 |

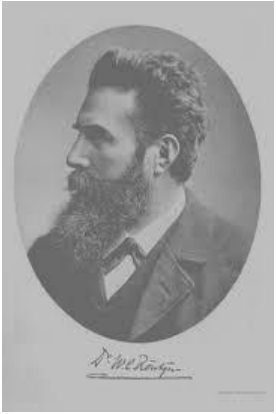
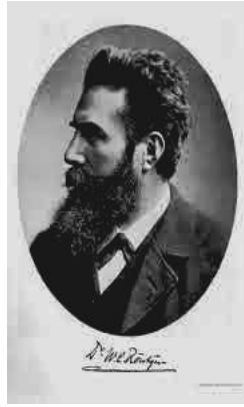
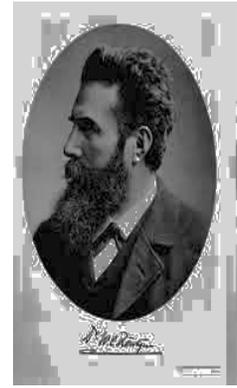

*Fig. 4. Image1 : original image*  *Fig. 5. HE enhanced image*  *Fig. 6. Fuzzy enhanced image*

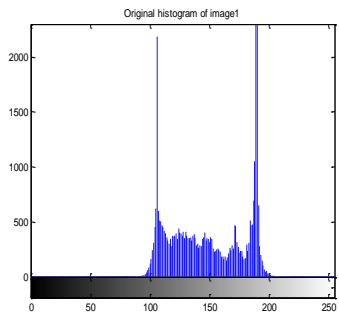
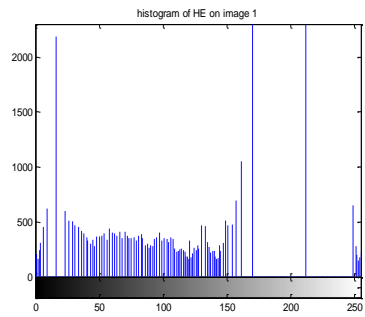
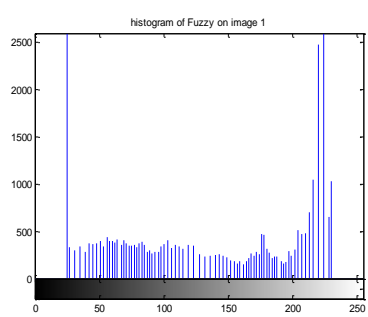

*Fig. 7. histogram of fig.4*  *Fig. 8. histogram of Fig.5*  *Fig. 9. histogram of Fig.6*

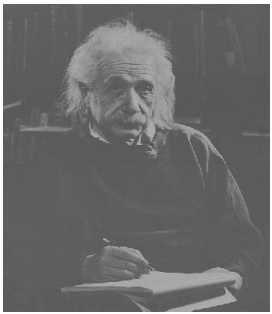
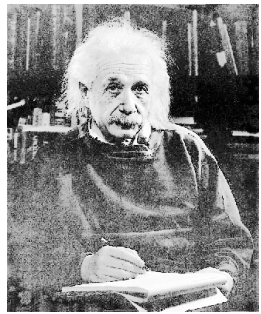
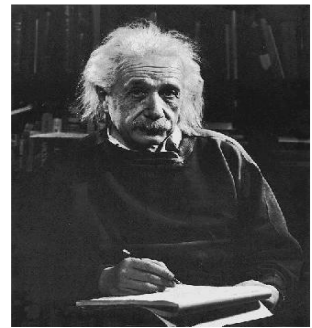

*Fig. 10. Image2: original image*  *Fig. 12. : HE enhanced image*  *Fig. 13.:Fuzzy enhanced image*

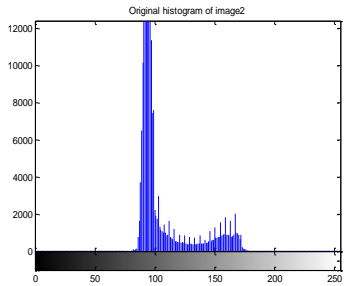 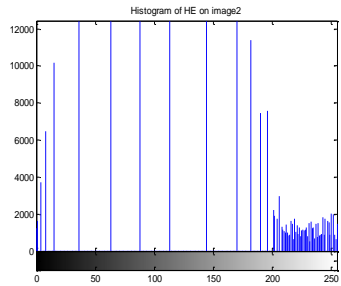 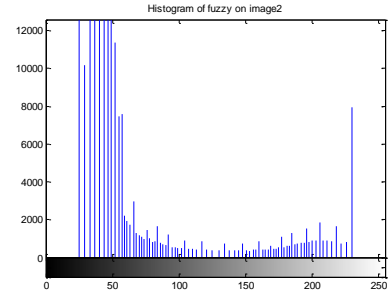

*Fig. 14: histogram of fig.10*         *Fig. 15: histogram of Fig.11*         *Fig. 16: histogram of Fig.13*

Using two low contrast images: image1 (figure4) and image2 (figure5) as shown above; concentration of pixels in the middle of the original histogram of image1 and image 2 is an indication that they are of low contrast. Pixels in the histogram of image 1 using HE(fig.5) are not evenly distributed over the gray scale unlike the use of fuzzy(fig.6), thus characteristics of the original image are more preserved with the fuzzy technique. Similarly, histogram of image 2. with the use of HE(fig.15) has its bright levels being moved right, thereby brightening the image more when compared with the use of fuzzy(fig.16).

## 4. CONCLUSIONS

Experimental results show that the application of fuzzy technique to image1(figure4) produced an image that contains lower mean squared error, higher peak to signal ratio, higher entropy and higher absolute mean brightness error than the image of HE. Thus output image from fuzzy results in less error, constructed image of higher quality, higher image details preservation and better brightness preservation than the image from HE. Again, for image2, output image of the fuzzy technique is of less error, higher image quality, higher image details preservation but lower brightness preservation than the image of the HE. Visually, it can be seen that some artifacts were introduced in image 1 when HE was applied and some details in image2 like details on the fore head were also lost in image2 with the use of HE. However, brightness preservation in image2 with the use of HE was higher than that obtained with fuzzy technique. Thus no technique is perfectly suitable for all enhancement cases but each technique is subject to the need of the system in question.

## REFERENCES

[1] Gonzalez R. and Woods R.: Digital Image Processing. (3rd ed.) Prentice hall, 2008.

[2] Deepika B., Renu B., and Vinod S. "Digital Image Enhancement by Improving Contrast, Removal of Noise and Motion Blurring", International Journal of Innovative Research in Science, Engineering and technology, vol 4, No.4, 2015, pp. 2601 2606. DOI: 10.15680/IJIRSET.2015.0404094

[3] Sapana S. and Vijaya K.. "Use of Histogram Equalization In Image Processing For Image Enhancement", International Journal of Software Engineering Research & Practices vol1, No.2, 2011

[4] Jaspreet K. and Amandeep K.. "Image Contrast Enhancement method based on Fuzzy Logic and Histogram Equalization", International Research Journal of Engineering and Technology (IRJET), vol3, No.5,2016, pp. 3089-3096.

[5] Kim Y. , "Contrast enhancement using brightness preserving bi-histogram equalization", IEEE Transactions On Consumer Electronics, vol 43, No. 1, 1997, pp.1-8.

[6] Soong-Der C. and Ramli A. "Minimum Mean brightness error bi-histogram equalization in contrast enhancement", IEEE Transactions on Consumer Electronics vol 49, No.4, 2003, pp. 1310 – 1319. DOI: 10.1109/TCE.2003.1261234

[7] Shefali G. and Yadwinder K.. "Review of Different Local and Global Contrast Enhancement Techniques for a Digital Image", International Journal of Computer Applications, vol.100, No.18, 2014, pp. 18 – 23.

[8] Reshmalakshmi C. Sasikumar M. and Shiny G. "Fuzzy Transform for Low-contrast Image Enhancement", International Journal of Applied Engineering Research, vol.13, No.11,2018, pp. 9103–9108

[9] Mamoria P. and Raj D. , "An Optimized Multiple Fuzzy Membership Functions based Image Contrast Enhancement Technique": KSII Transactions on Internet & Information Systems, vol. 12, No.3,2018, pp.1206-1223.

[10] Tizhoosh, R. and HauBecker H . "Fuzzy Image Processing", Handbook of Computer Vision and Applications, vol.2, 1999, pp. 549-556. DOI: 10.1016/B978-012379777-3/50017-0.

[11] Sundaram M., Ramar K., Arumugam N. and Prabin G. (2011). "Histogram Modified Local Contrast Enhancement for Mammogram Images", Applied Soft Computing, vol.11, No.8,2011,pp. 5809-5816.

[12] Vinod K. and Rahul R., "A Comparative Analysis of Image Contrast Enhancement Techniques based on Histogram Equalization for Gray Scale Static Images", International Journal of Computer Applications, vol.45, No.21, 2012

[13] Raju S. and Sawant B. "Noisy Fingerprint Image Enhancement Technique for Image Analysis: A Structure Similarity Measure Approach", International Journal of Computer Science and Network Security (IJCSNS), vol.7, No.9.,2007